\documentclass{article}

\usepackage{microtype}
\usepackage{graphicx}
\usepackage{subfigure}
\usepackage{booktabs} 

\usepackage{hyperref}

\usepackage[accepted]{icml2025}

\usepackage{amsmath}
\usepackage{amssymb}
\usepackage{mathtools}
\usepackage{amsthm}
\usepackage{multirow}
\usepackage{paralist} 
\usepackage{tabularx}

\usepackage[capitalize,noabbrev]{cleveref}

\theoremstyle{plain}

\theoremstyle{definition}

\theoremstyle{remark}

\usepackage[textsize=tiny]{todonotes}
\definecolor{myYellow}{rgb}{0.8, 0.6, 0.1} 

\icmltitlerunning{Submission and Formatting Instructions for ICML 2025}

\begin{document}

\twocolumn[
\icmltitle{Harmonizing Intra-coherence and Inter-divergence in Ensemble Attacks for Adversarial Transferability}



\icmlsetsymbol{equal}{*}

\begin{icmlauthorlist}
\icmlauthor{Zhaoyang Ma}{bjtu}
\icmlauthor{Zhihao Wu}{bjtu}
\icmlauthor{Wang Lu}{thu}
\icmlauthor{Xin Gao}{pku}
\icmlauthor{Jinghang Yue}{bjtu}
\icmlauthor{Taolin Zhang}{thu}
\icmlauthor{Lipo Wang}{ntu}
\icmlauthor{Youfang Lin}{bjtu}
\icmlauthor{Jing Wang}{bjtu}
\end{icmlauthorlist}

\icmlaffiliation{bjtu}{School of Computer Science & Technology, Beijing, China}
\icmlaffiliation{thu}{Tsinghua University, Beijing, China}
\icmlaffiliation{pku}{Institute for AI, Peking University, Beijing, China}
\icmlaffiliation{ntu}{School of Electrical and Electronic Engineering, Nanyang Technological University, Singapore}

\icmlcorrespondingauthor{Jing Wang}{wj@bjtu.edu.cn}
\icmlcorrespondingauthor{Wang Lu}{luw12@tsinghua.org.cn}

\icmlkeywords{Machine Learning, ICML}

\vskip 0.3in
]




\begin{abstract}
The development of model ensemble attacks has significantly improved the transferability of adversarial examples, but this progress also poses severe threats to the security of deep neural networks. Existing methods, however, face two critical challenges: insufficient capture of shared gradient directions across models and a lack of adaptive weight allocation mechanisms. To address these issues, we propose a novel method —\textbf{H}armonized \textbf{E}nsemble for \textbf{A}dversarial \textbf{T}ransferability (\textbf{HEAT})—which introduces domain generalization into adversarial example generation for the first time. 
HEAT consists of two key modules: Consensus Gradient Direction Synthesizer, which uses Singular Value Decomposition to synthesize shared gradient directions; and Dual-Harmony Weight Orchestrator which dynamically balances intra-domain coherence, stabilizing gradients within individual models, and inter-domain diversity, enhancing transferability across models.
Experimental results demonstrate that HEAT significantly outperforms existing methods across various datasets and settings, offering a new perspective and direction for adversarial attack research.
\end{abstract}

\section{Introdution}
\label{introduction}

\begin{figure}[ht]
\centering
\includegraphics[width=1\linewidth]{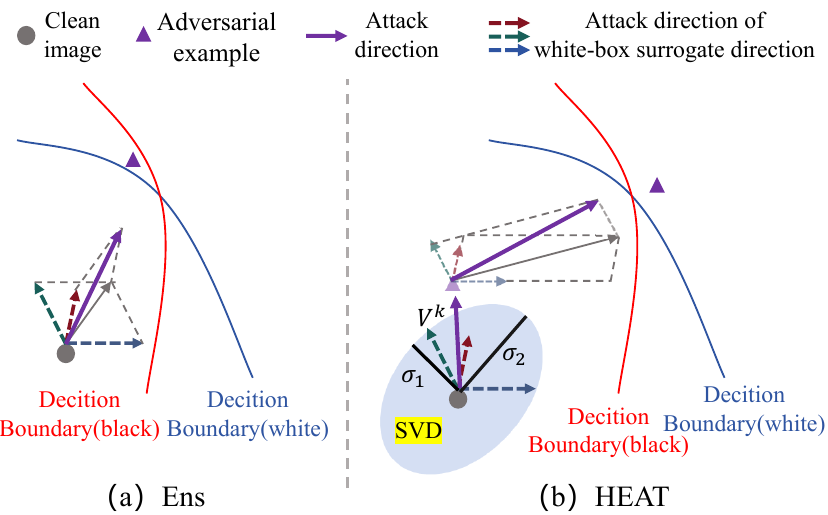}
\caption{(a) Ens averages gradients from multiple surrogate models to determine the attack direction, but its transferability is limited due to the absence of dynamic weight adjustment. (b) HEAT synthesizes a common gradient direction and dynamically adjusts the weights of gradients from different surrogate models, significantly improving the attack transferability.}
\label{fig:motivation}
\end{figure}

Deep learning has achieved remarkable progress, with applications in autonomous driving \cite{gao2024ethical}, disease diagnosis \cite{ma2023homologous}, and intelligent surveillance \cite{pazho2023ancilia}. However, recent deep neural networks (DNNs) are vulnerable to adversarial examples—inputs perturbed by imperceptible modifications that frequently lead to incorrect predictions \cite{szegedy2013intriguing}. More concerning is the transferability of adversarial examples; samples designed for one model often mislead others \cite{goodfellow2014explaining}. This allows attackers to carry out effective attacks without prior knowledge of the target model’s architecture or parameters, posing a serious threat to black-box DNN applications.

Recent research has focused on enhancing the transferability of adversarial examples to improve the robustness of DNNs, offering deeper insights into their internal mechanisms and bolstering model defenses. Adversarial attacks are categorized into three types: input transformation, gradient enhancement, and model ensemble attacks. Among these, model ensemble attacks show superior success rates in black-box scenarios compared to single-model approaches \cite{liu2022delving}. This advantage arises from the ability of ensemble methods to integrate gradient information from multiple surrogate models during adversarial example generation. By doing so, they reduce the risk of overfitting to the specific characteristics of a single model, thereby increasing the likelihood of success against diverse target models.

Despite the success of ensemble attack methods in generating universal adversarial perturbations, they struggle to capture common gradient directions and lack adaptive weighting mechanisms. The non-convex nature of DNNs, high-dimensional parameter spaces, and randomness in model architectures and initializations often lead to significant divergence in the gradient directions of different models when processing the same input \cite{li2018visualizing}. Moreover, existing methods typically aggregate gradients using fixed weights or simple averaging \cite{tramer2018ensemble, linnesterov}, lacking dynamic adjustment based on adversarial effectiveness. This underestimates the contribution of models exhibiting higher adversarial efficacy, diminishing attack effectiveness in multi-model scenarios.

In this paper, we propose a novel method, \underline{H}armonized \underline{E}nsemble for \underline{A}dversarial \underline{T}ransferability (\textbf{HEAT}), integrating domain generalization into the design of model ensemble attacks. To our knowledge, this is the first work to incorporate domain generalization principles into ensemble attack design, offering a fresh approach that enhances the adaptability of adversarial examples across diverse models. Figure \ref{fig:motivation} illustrates how HEAT advances beyond conventional ensemble approaches with two specialized modules: \underline{C}onsensus \underline{Gra}dient \underline{D}irection \underline{S}ynthesizer (\textbf{C-GRADS}) and \underline{D}ual-\underline{Harm}ony Weight \underline{O}rchestrator (\textbf{D-HARMO}). 

C-GRADS employs Singular Value Decomposition (SVD) to synthesize shared gradient directions across surrogate models,  enabling efficient perturbations that robustly transfer to diverse black-box models. D-HARMO implements dual adaptive weighting: (1) \textbf{Intra-domain coherence} quantifies the stability of gradient directions within individual models to ensure adversarial perturbations effectively exploit vulnerabilities, prioritizing models with greater breadth; (2) \textbf{Inter-domain divergence} leverages the diversity of gradient directions across multiple models to improve the transferability of adversarial examples to unseen targets, prioritizing models with greater depth. D-HARMO act as a strategic team builder: intra-domain coherence weights "generalists" for broad robustness, while inter-domain divergence prioritizes "specialists" with transfer-critical features, balancing attack breadth and depth. Comprehensive evaluations on CIFAR-10, CIFAR-100 and ImageNet benchmarks demonstrating that HEAT substantially outperforms existing methods across various test settings.

The main contributions are summarized as follows:
\begin{itemize}
    \item We establish the first formal connection between domain generalization and adversarial attacks, pioneering an approach that improves the adaptability of adversarial examples across diverse models and provides a novel perspective for existing research.
    \item We propose  Harmonized Ensemble for Adversarial Transferability (HEAT), which substantially enhances the transferability of adversarial examples by harmonizing gradient information across multiple models.
    \item HEAT redefines ensemble attacks from the domain generalization perspective: C-GRADS synthesizes shared gradient directions across models, while D-HARMO integrates intra-domain coherence and inter-domain divergence, optimizing gradient information utilization.
    \item The proposed HEAT generates adversarial examples with high transferability, significantly outperforming existing ensemble attack methods in terms of effectiveness and adaptability.
\end{itemize}

\begin{figure*}[ht]
\centering
\includegraphics[width=1\linewidth]{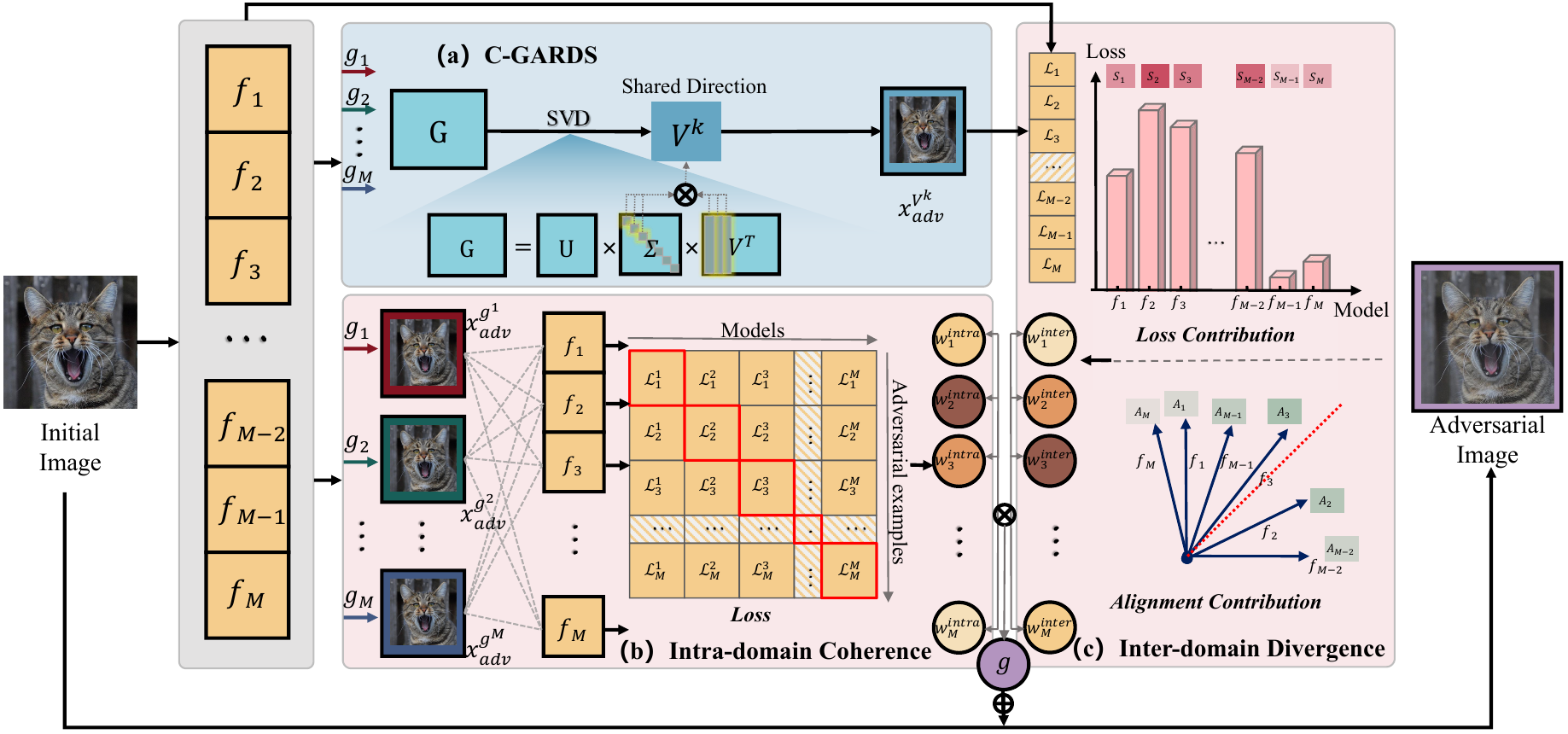}
\caption{An overview of HEAT: Components (b) and (c) collectively form the D-HARMO module. In (a), we use SVD to synthesize shared gradient directions; (b) captures synergistic contributions to ensure consistency in gradient directions across models; and (c) dynamically adjusts weights based on loss and alignment contributions among the models. The refined gradients generated by these components are integrated to produce highly effective adversarial images.}
\label{fig:HEAT}
\end{figure*}

\section{Related work}

Adversarial attacks were first formalized by \citet{szegedy2013intriguing}, showing that imperceptible perturbations to input data could cause significant misclassifications in DNNs. This discovery exposed critical vulnerabilities in machine learning systems, sparking extensive research into adversarial attacks. 

\textbf{Input transformation attacks} manipulate input data through geometric deformations, pixel-level perturbations, or structured noise injection, thereby misleading model predictions. Common approaches include: gradient-based perturbation \cite{goodfellow2014explaining,carlini2017towards}, random/structural transformations \cite{xie2019improving,wang2023structure}, reinforcement learning-driven dynamic transformations \cite{zhu2024learning}, and block-wise transformations \cite{wang2024boosting}.

\textbf{Gradient enhancement attacks} optimize gradient computations to craft transfer-robust adversarial examples. This approach improves attack effectiveness through more precise gradient calculations. Common approaches include: update direction optimization \cite{zhu2023boosting}, momentum-based smoothing \cite{dong2018boosting}, adaptive iterative optimization \cite{li2023adaptive}, cross-domain fusion \cite{pang2024spatial,tao2024ags}, and self-enhancement mechanisms\cite{wu2024improving}.

\textbf{Model ensemble attacks} integrate gradients from multiple surrogate models to enhance cross-model transferability, addressing the overfitting limitations of single-model attacks. Some ensemble approaches include longitudinal strategy \cite{li2020learning}, gradient fusion \cite{liu2022delving}, Bayesian ensemble optimization \cite{li2023making}, loss landscape flattening \cite{chen2023rethinking}, relational graph-based \cite{pi2023improving}, and dynamic output fusion \cite{chen2023adaptive}.

\section{Methodology}

\subsection{General Overview}

In this study, for a given input image $x$, we iteratively generate an adversarial example $x_{adv}^i$ by updating the perturbation $\sigma_i$. The process initializes with $x_{adv}^1 = x + \sigma_1$, where $\sigma_1$ denotes the initial random perturbation. Through iterative refinement, each perturbation manipulates the model's decision boundaries to induce misclassification. To keep the perturbation imperceptible, we enforce an $\ell_p$-norm constraint on its magnitude. The optimization terminates upon convergence, yielding a final adversarial example $x_{adv}^{T}$ that maximally disrupts the model's predictions while preserving the semantic content of the original image. 

Adversarial perturbation generation is formulated as a constrained optimization problem:
\begin{equation}
\min_{\delta} \mathcal{L}(f(x + \delta), y), \quad \text{subject to} \quad \|\delta\|_p \leq \epsilon,
\end{equation}
where $\mathcal{L}(\cdot)$ denotes the cross-entropy loss function quantifying the discrepancy between model predictions $f(\cdot)$ and ground truth label $y$. The $\ell_p$-norm constraint ($p \in \{2, \infty\}$) regulates the perturbation's perceptibility, with $\varepsilon$ controlling the maximum allowable distortion. 

To solve this optimization problem, we employ gradient-based optimization. 
Starting from an initial adversarial example, the method iteratively refines the perturbation using gradient information:
\begin{equation}
x_{\text{adv}}^i = \text{Clip}_\epsilon \left( x_{\text{adv}}^{i-1} + \alpha \cdot \text{sign} \left( \nabla_x L(f(x_{\text{adv}}^{i-1}), y) \right) \right),
\end{equation}
where $\mathrm{Clip}_\epsilon(\cdot)$ projects the perturbation back to the $\epsilon$-ball constraint, and $\alpha$ controls the step size. This iterative process ensures the adversarial example remains within the distortion bound while maximizing the model's loss.

To enhance the transferability of adversarial examples across different models, we implement an ensemble attack strategy that aggregates gradient from multiple surrogate models:
\begin{equation}
x_{adv}^i = \mathrm{Clip}_\epsilon \Bigg( x_{adv}^{i-1} + \alpha \cdot \mathrm{sign} \Bigg( \frac{1}{M} \sum_{m=1}^M \nabla_x \mathcal{L}(f_m(x_{adv}^{i-1}), y) \Bigg) \Bigg),
\end{equation}
where $M$ is the number of surrogate models. By averaging gradients from multiple models, this strategy enhances perturbation robustness and generalizability, ensuring strong attack performance across diverse target models.

\subsection{Harmonized Ensemble for Adversarial Transferability}

In this work, we propose HEAT, an innovative method for generating highly transferable adversarial examples, inspired by domain generalization to enhance transferability through model ensemble attacks. HEAT comprises two key modules: Consensus Gradient Direction Synthesizer (\textbf{C-GRADS}) and Dual-HarMony Weight Orchestrator (\textbf{D-HARMO}). C-GRADS synthesizes shared gradient directions by aggregating the most influential perturbation directions across multiple models, improving adversarial example transferability. D-HARMO dynamically allocates weights to each model's gradient contributions by balancing intra-domain coherence and inter-domain divergence, prioritizing models with greater breadth and depth. An overview of HEAT is shown in Figure \ref{fig:HEAT}, while the algorithmic workflow is presented in Algorithm \ref{alg:heat}.

\subsubsection{Consensus Gradient Direction Synthesizer}

C-GRADS enhances adversarial transferability by synthesizing shared gradient directions across models. Different models often exhibit significant variations in gradient directions for the same input, undermining the transferability of adversarial samples and limiting conventional adversarial generation methods. To tackle this challenge, we propose the C-GRADS module.

Specifically, given an input sample $x$, we apply a small random perturbation to generate the initial adversarial example $x_{\text{adv}}^\text{init}$. For $x_{\text{adv}}^\text{init}$, we compute the gradients $\nabla_{x} \mathcal{L}m(x{\text{adv}}^\text{init})$ for the loss functions of $M$ white-box models, where $m = 1, 2, \dots, M$. These gradients quantify the sensitivity of each model's loss function to perturbations in $x_{\text{adv}}^\text{init}$, revealing potential optimization directions in the input space.

To extract shared gradient directions efficiently, we construct a gradient matrix $\mathbf{G} \in \mathbb{R}^{M \times D}$ by stacking the gradient vectors row-wise, where each row corresponds to the gradient vector of a model, and $D$ denotes the input dimensionality:
\begin{equation}
\mathbf{G} = 
\begin{bmatrix}
\text{vec}(\nabla_{x} \mathcal{L}_1(x_{\text{adv}}^\text{init})) \\
\text{vec}(\nabla_{x} \mathcal{L}_2(x_{\text{adv}}^\text{init})) \\
\vdots \\
\text{vec}(\nabla_{x} \mathcal{L}_M(x_{\text{adv}}^\text{init}))
\end{bmatrix},
\end{equation}

However, directly using these gradients is often ineffective due to significant discrepancies in gradient directions across models, limiting adversarial example transferability. To enhance adversarial transferability, we focus on identifying common sensitive directions that are shared across models. We identify shared sensitive directions by performing Singular Value Decomposition (SVD) \cite{eckart1936approximation} on the gradient matrix $G$, decomposing it into:
\begin{equation}
\label{eq:svd}
\mathbf{G} = \mathbf{U} \mathbf{\Sigma} \mathbf{V}^\top,
\end{equation}
where the singular value matrix $\Sigma$ contains singular values $\sigma_1, \sigma_2, \dots, \sigma_M$. Arranged in descending order ($\sigma_1 \geq \sigma_2 \geq \dots \geq \sigma_M$), these singular values quantify the contribution of each principal direction, reflecting their relative importance. The columns of the right singular vector matrix $V$, $v_1, v_2, \dots, v_D$, represent the principal directions in the input feature space. The importance of these column vectors decreases from left to right, as determined by their corresponding singular values, with $\sigma_1$ being most strongly associated with the direction $v_1$, and so on.

To dynamically identify the most critical gradient directions, we calculate the contribution of each singular value $\sigma_i$ to the total variance.Based on a predefined cumulative contribution ratio $p$, we determine the number of singular vectors $k$ to retain. This can be formally expressed as:
\begin{equation}
\label{eq:k}
    k = \min \left\{ k \, \middle| \, \frac{\sum_{i=1}^{k} \sigma_i}{\sum_{i=1}^{M} \sigma_i} \geq p \right\},
\end{equation}
where $k$ is the smallest integer satisfying the inequality. This $k$ ensures that the first $k$ singular values capture at least $p$ of the total variance. By selecting the top $k$ singular vectors, we capture the most representative common gradient directions, which significantly contribute to $G$. These directions effectively encode the primary sensitivities of the models.

Next, we compute the shared gradient direction $\mathbf{V}_k$ using the selected $k$ singular vectors. We compute a weighted sum of the selected singular vectors, with weights proportional to their corresponding singular values:
\begin{equation}
\label{eq:vk}
\mathbf{V}_k = \sum_{i=1}^{k} \sigma_i \mathbf{v}_i.
\end{equation}
Using $\mathbf{V}_k$, we generate the adversarial example $x_{\text{adv}}^{\mathbf{V}_k}$:
\begin{equation}
\label{eq:advvk}
x_{\text{adv}}^{V_k} = \text{Clip}_\epsilon \left( x + \alpha \cdot \text{sign}(\mathbf{V_k}) \right),
\end{equation}
where $\alpha$ controls the perturbation magnitude, and $\text{sign}(\mathbf{V}_k)$ indicates the perturbation direction along $\mathbf{V}_k$. Perturbations along these shared directions enhance adversarial example transferability across models, significantly improving attack effectiveness.

\subsubsection{Dual-Harmony Weight Orchestrator}
\label{D-HARMO}

In model ensemble attacks, traditional methods often use fixed weights or simple gradient averaging to aggregate gradients from multiple models. Although straightforward, these static methods ignore the unique characteristics and dynamic contributions of individual models during adversarial example generation. Consequently, they neglect critical gradient interactions between models and fail to leverage their complementary strengths, limiting adversarial example effectiveness and transferability.

To overcome these limitations, we propose a novel dynamic weight allocation strategy specifically designed for ensemble attacks. Drawing inspiration from domain generalization, we model gradient interactions with two complementary components: \textbf{intra-domain coherence} and \textbf{inter-domain divergence}. Intra-domain coherence measures the consistency of gradient directions within white-box models. Conversely, inter-domain divergence leverages gradient discrepancies between models, dynamically adjusting weights based on their sensitivity and alignment. Together, these components form a unified and adaptive weight allocation framework, significantly enhancing adversarial example diversity, transferability, and effectiveness.

\textbf{Intra-domain Coherence}

Intra-domain coherence models the consistency of gradient directions across models, to those with strong gradient consistency, reflecting their extensive breadth. These models function like versatile team players, excelling in coordination to stabilize and improve adversarial example generation. Unlike static directions derived from C-GRADS, this module computes weights in real-time, dynamically adapting to variations across models during adversarial example generation, enhancing ensemble attack effectiveness.

Specifically, for $M$ white-box models, the loss function for each model $m$ is defined as $\mathcal{L}m(x{\text{adv}}^{\text{init}})$. Using the gradient direction of each model, we generate the corresponding adversarial example as follows:
\begin{equation}
\label{eq:advinitm}
x_{\text{adv}}^{\text{init},m} = x + \alpha \cdot \text{sign}(\nabla_x \mathcal{L}_m(x_{\text{adv}}^{\text{init}})), \quad m \in \{1, 2, \dots, M\}. 
\end{equation}
For each adversarial example $x_{\text{adv}}^{\text{init},m}$ generated by model $m$, we calculate its self-prediction error as follows:
\begin{equation}
\label{eq:selfloss}
\mathcal{L}_m^{\text{self}} = \mathcal{L}(f_m(x_{\text{adv}}^{\text{init},m}), y),  
\end{equation}
Here, $\mathcal{L}(\cdot)
$ is the loss function, typically the cross-entropy; $f_m(x_{\text{adv}}^{\text{init},m})$ denotes the model $m$'s prediction for the adversarial example $x_{\text{adv}}^{\text{init},m}$; and $y$ refers to the ground truth label. 

To quantify the relative importance of each model during adversarial example generation, we calculate the intra-domain dynamic weight $w_m^{\text{intra}}$ for each model $m$. For a given model $m$ and other models $j$, $w_m^{\text{intra}}$ is computed as:
\begin{align}
\label{eq:wintra1}
w_m^{\text{intra}} &= \sum_{j \neq m, j=1}^M 
\frac{\log\big(\mathcal{L}(f_j(x_{\text{adv}}^{\text{init},m}), y) + \epsilon\big)}
{\mathcal{L}_j^{\text{self}} + \epsilon}, \nonumber \\
&\quad m \in \{1, 2, \dots, M\},
\end{align}
Here, $\epsilon$ is a small constant to prevent division by zero. The intra-domain weights $\mathbf{w}^{\text{intra}}$ are normalized across models:
\begin{equation}
\label{eq:wintra2}
\tilde{w}_m^{\text{intra}} = \frac{w_m^{\text{intra}}}{\sum_{k=1}^M w_k^{\text{intra}}}, \quad m \in \{1, 2, \dots, M\}.
\end{equation}
This process generates the intra-domain weights for all white-box models participating in the ensemble attack:
\begin{equation}
\label{eq:wintra3}
{w}^{\text{intra}} = [\tilde{w}_1^{\text{intra}},\tilde{w}_2^{\text{intra}}, \dots, \tilde{w}_M^{\text{intra}}].
\end{equation}
These weights capture the synergy of gradient directions, coordinate gradient contributions among models, and generate a consistent and efficient shared sensitive direction, enhancing the effectiveness of ensemble attacks.

\textbf{Inter-domain Divergence}

Inter-domain divergence captures variations in gradient directions, magnitudes, and adversarial sensitivities across models, assigning higher weights to those with greater depth, analogous to specialists excelling in their respective domains within a team. Unlike intra-domain coherence, which enhances gradient coordination within models, inter-domain divergence identifies and leverage differences across them.

Inter-domain divergence is modeled using the \textbf{Loss Contribution Factor $\mathcal{S}$} and the \textbf{Alignment Contribution Factor $\mathcal{A}$}, which quantify a model’s sensitivity to the shared adversarial example $x_{\text{adv}}^{V_k}$ and its gradient alignment with other models. This approach captures the differences among models, enabling precise allocation of inter-domain weights.

The loss contribution factor $\mathcal{S}_m$ measures the sensitivity of each model $m$ to a given adversarial example. The loss contribution factor $\mathcal{S}_m$ for each model is defined as:
\begin{equation}
\label{eq:losscontribution}
\mathcal{S}_m = \mathcal{L}_m(x_{\text{adv}}^{V_k}) + \epsilon, \quad m \in \{1, 2, \dots, M\}.
\end{equation}
We define the alignment contribution factor $\mathcal{A}_m$, quantifying the alignment of gradient directions among models, assessing each model's representativeness and influence. The alignment contribution factor $\mathcal{A}_m$ is computed by evaluating the gradient similarity $P_{mj}$ between model $m$ and each model $j$, using cosine similarity:
\begin{equation}
\label{eq:cos}
P_{mj} = \frac{\nabla_{x} \mathcal{L}_m(x_{\text{adv}}^{V_k}) \cdot \nabla_{x} \mathcal{L}_j(x_{\text{adv}}^{V_k})}{\|\nabla_{x} \mathcal{L}_m(x_{\text{adv}}^{V_k})\| \|\nabla_{x} \mathcal{L}_j(x_{\text{adv}}^{V_k})\|}.
\end{equation}
Subsequently, $\mathcal{A}_m$ for model $m$ is derived by aggregating its gradient similarity with all other models:
\begin{equation}
\label{eq:aligncontribution}
\mathcal{A}_m = \frac{1}{\frac{1}{M-1} \sum_{j \neq m} P_{mj} + \epsilon}.
\end{equation}
Models with higher gradient similarity (i.e., greater alignment) are assigned larger alignment contribution factors, increasing their influence on the final gradient computation. 

To ensure $\mathcal{S}_m$ and $\mathcal{A}_i$ are comparable, we apply normalization. A temperature parameter $\tau$ is introduced to control the smoothness of the normalization results:
\begin{equation}
\label{eq:norms}
\tilde{\mathcal{S}}_m = \left( \frac{\mathcal{S}_m}{\sum_{k=1}^{M} \mathcal{S}_k} \right)^{\frac{1}{\tau}},
\end{equation}
\begin{equation}
\label{eq:norma}
\tilde{\mathcal{A}}_m = \left( \frac{\mathcal{A}_m}{\sum_{k=1}^{M} \mathcal{A}_k} \right)^{\frac{1}{\tau}}.
\end{equation}

We define the information entropy $\mathcal{H}_m$ to quantify the uncertainty in each model's contribution. The information entropy captures the uncertainty of model $m$ by considering its loss and alignment contributions. Models with substantial loss and alignment contributions exhibit low information entropy, indicating high confidence in their gradient contributions. Conversely, higher information entropy suggests greater uncertainty in the model's gradient contribution. The information entropy is calculated as follows:
\begin{equation}
\label{eq:H}
\mathcal{H}_m = -\left( \tilde{\mathcal{S}}_m \log \tilde{\mathcal{S}}_m + \tilde{\mathcal{A}}_m \log \tilde{\mathcal{A}}_m \right).
\end{equation}
The inter-domain weight $w_m^{\text{inter}}$ for model $m$ is defined as the reciprocal of its information entropy:
\begin{equation}
\label{eq:winter1}
w_m^{\text{inter}} = \frac{1}{\mathcal{H}_m + \epsilon}.
\end{equation}

The inter-domain weights $w_m^{\text{inter}}$ are normalized as follows:
\begin{equation}
\label{eq:winter2}
\tilde{w}_m^{\text{inter}} = \frac{w_m^{\text{inter}}}{\sum_{k=1}^{M} w_k^{\text{inter}}}.
\end{equation}

This process generates the inter-domain weights for all white-box surrogate models in the ensemble attack:
\begin{equation}
\label{eq:winter3}
w^{\text{inter}} = [\tilde{w}_1^{\text{inter}}, \tilde{w}_2^{\text{inter}}, \dots, \tilde{w}_M^{\text{inter}}].
\end{equation}
These weights capture the dynamic relationships in inter-domain gradient discrepancies, facilitating adaptive weight allocation that leverages each model's unique gradient. This enhances the transferability and diversity of adversarial examples, improving the effectiveness of ensemble attacks.

\textbf{Adversarial Example Generation}

The final adversarial gradient $\mathbf{g}$ integrates the intra-domain weight $w^{\text{intra}}$ and inter-domain weight $\mathbf{w}^{\text{inter}}$, formulated as:
\begin{equation}
\label{eq:gradient}
\mathbf{g} = \sum_{m=1}^{M} w_m^{\text{intra}} \cdot w_m^{\text{inter}} \cdot \nabla_x \mathcal{L}_m(x_{\text{adv}}^{V_k}). 
\end{equation}

Utilizing the gradient $\mathbf{g}$, the adversarial example is updated through the following formulation:
\begin{equation}
\label{eq:final}
x_{\text{adv}}^* = \text{Clip}_\epsilon \left( x + \alpha \cdot \text{sign}(\mathbf{g}) \right).
\end{equation}
This update ensures that the adversarial example exploits the vulnerabilities of multiple models while maintaining its transferability across black-box scenarios.

\section{Experiments}
\subsection{Experimental Setting}
\textbf{Datasets.} The experiments are performed on CIFAR-10, CIFAR-100, and ImageNet \cite{krizhevsky2009learning,deng2009imagenet}, which are standard benchmarks for classification and adversarial attack evaluation \cite{dong2020benchmarking}.

\textbf{Comparision.}
We compare HEAT with Ens \cite{liu2022delving}, SVRE \cite{xiong2022stochastic}, and AdaEA \cite{chen2023adaptive} using the same implementation.

\newpage
\begin{algorithm}[H]
\caption{The HEAT algorithm}
\label{alg:heat}
\begin{algorithmic}[1]
\INPUT Input sample $x$, ensemble of $M$ white-box models $\{f_m\}_{m=1}^M$, perturbation magnitude $\alpha$, cumulative contribution ratio $p$, temperature parameter $\tau$, and small constant $\epsilon$
\OUTPUT Adversarial example $x_{\text{adv}}^{*}$

\STATE $x_{\text{adv}}^\text{init} \gets x$

\STATE \textcolor{myYellow}{\# C-GRADS}
\STATE Compute gradients $\{\nabla_{x} \mathcal{L}_m(x_{\text{adv}}^\text{init})\}_{m=1}^M$ for all models
\STATE Construct gradient matrix $\mathbf{G} \in \mathbb{R}^{M \times D}$ by stacking gradients row by row
\STATE Perform SVD on $\mathbf{G}$ using Eq. \ref{eq:svd}
\STATE Determine the number of singular vectors $k$ based on cumulative contribution ratio $p$ using Eq. \ref{eq:k}
\STATE Compute shared gradient direction $\mathbf{V}_k$ using Eq. \ref{eq:vk}
\STATE Generate adversarial example $x_{\text{adv}}^{\mathbf{V}_k}$ using Eq. \ref{eq:advvk} 

\STATE \textcolor{myYellow}{\# Intra-domain Coherence}

\FOR{$m \leftarrow 1$ to $M$}
    \STATE Compute adversarial example $x_{\text{adv}}^{\text{init},m}$ for model $m$ using Eq. \ref{eq:advinitm}
    \STATE Compute the self-prediction loss $\mathcal{L}_m^{\text{self}}$ using Eq. \ref{eq:selfloss}
    \STATE Compute intra-domain weight $w_m^{\text{intra}}$ using Eq. \ref{eq:wintra1}
\ENDFOR
\STATE Normalize intra-domain weights $\{w_m^{\text{intra}}\}_{m=1}^M$ Eq. \ref{eq:wintra2}
\STATE Compute intra-domain weight $w^{\text{intra}}$ using Eq.\ref{eq:wintra3}

\STATE \textcolor{myYellow}{\# Inter-domain Divergence}
\FOR{$m \leftarrow 1$ to $M$}
    \STATE Compute loss contribution factor $\mathcal{S}_m$ using Eq. \ref{eq:losscontribution}
    \STATE Compute alignment contribution factor $\mathcal{A}_m$ using Eq. \ref{eq:cos}–\ref{eq:aligncontribution}
    \STATE Normalize $\mathcal{S}_m$ and $\mathcal{A}_m$ using Eq. \ref{eq:norms}–\ref{eq:norma}
    \STATE Compute information entropy $\mathcal{H}_m$ using Eq. \ref{eq:H}
    \STATE Compute inter-domain weight $w_m^{\text{inter}}$ using Eq. \ref{eq:winter1}
\ENDFOR
\STATE Normalize inter-domain weights $\{w_m^{\text{inter}}\}_{m=1}^M$ using Eq. \ref{eq:winter2}
\STATE Compute inter-domain weight $w^{\text{inter}}$ using Eq. \ref{eq:winter3}

\STATE \textcolor{myYellow}{\# Adversarial Gradient Combination}
\STATE Compute the final adversarial gradient $\mathbf{g}$ using Eq. \ref{eq:gradient}

\STATE Update adversarial example $x_{\text{adv}}^*$ using Eq. \ref{eq:final}
\end{algorithmic}
\end{algorithm}

\begin{table*}[ht]
\centering
\begin{tabular}{cccccccccccc}
\hline
\textbf{Dataset} & \textbf{Method} & \textbf{Res-50} & \textbf{WRN101-2} & \textbf{BiT-50} & \textbf{BiT-101} & \textbf{ViT-B} & \textbf{DeiT-B} & \textbf{Swin-B} & \textbf{Swin-S} & \textbf{Average} \\
\hline
\multirow{4}{*}{CIFAR-10} & Ens & 17.20 & 22.56 & 13.72 & 12.32 & 6.83 & 10.49 & 7.93 & 12.63 & 12.96 \\
 & SVRE & 35.40 & 32.55 & 25.75 & 24.98 & 16.51 & 20.57 & 14.28 & 20.05 & 23.76 \\
 & AdaEA & 33.58 & 34.43 & 37.12 & 33.21 & 28.10 & 31.77 & 28.88 & 31.16 & 32.28 \\
 & HEAT & 39.17 & 32.58 & 33.96 & 35.51 & 38.18 & 51.57 & 40.36 & 51.12 & \textbf{40.31} \\
\hline
\multirow{4}{*}{CIFAR-100} & Ens & 65.00 & 69.94 & 52.00 & 55.62 & 39.30 & 53.33 & 39.91 & 51.21 & 53.29 \\
 & SVRE & 61.05 & 62.06 & 54.71 & 47.53 & 43.52 & 47.83 & 41.6 & 54.91 & 51.65 \\
 & AdaEA & 67.43 & 70.95 & 58.33 & 55.38 & 56.36 & 65.50 & 51.29 & 60.62 & 60.73 \\
 & HEAT & 68.13 & 61.31 & 56.33 & 46.90 & 55.63 & 70.78 & 62.35 & 71.64 & \textbf{61.63} \\
\hline
\multirow{4}{*}{ImageNet} & Ens & 22.86 & 27.40 & 24.80 & 21.80 & 11.67 & 16.88 & 6.84 & 14.68 & 18.37 \\
 & SVRE & 27.72 & 31.99 & 30.83 & 29.31 & 21.77 & 24.81 & 16.60 & 22.43 & 25.68 \\
 & AdaEA & 40.87 & 44.42 & 45.26 & 40.83 & 35.61 & 40.76 & 25.71 & 33.09 & 38.32 \\
 & HEAT & 45.72 & 48.65 & 50.65 & 45.16 & 42.72 & 52.98 & 27.25 & 39.71 & \textbf{44.11} \\
\hline
\end{tabular}
\caption{Attack success rate (\%) of black-box models on CIFAR-10, CIFAR-100, and ImageNet using different ensemble attack methods.}
\label{tab:performance}
\end{table*}

\begin{table*}[ht]
\centering
\begin{tabular}{cccccccccccc}
\hline
\textbf{Base} & \textbf{Attack} & \textbf{Res-50} & \textbf{WRN101-2} & \textbf{BiT-50} & \textbf{BiT-101} & \textbf{ViT-B} & \textbf{DeiT-B} & \textbf{Swin-B} & \textbf{Swin-T} & \textbf{Average} \\
\hline
\multirow{4}{*}{I-FGSM} & Ens & 30.11 & 15.90 & 13.44 & 10.06 & 7.90 & 20.04 & 15.44 & 22.09 & 16.87 \\
 & SVRE & 51.92 & 27.50 & 22.90 & 18.29 & 13.30 & 30.74 & 24.84 & 51.01 & 30.06 \\
 & AdaEA & 56.22 & 30.23 & 24.11 & 19.91 & 18.59 & 43.95 & 36.18 & 48.98 & 34.77 \\
 & HEAT & 46.47 & 33.75 & 37.95 & 34.65 & 43.40 & 60.12 & 47.35 & 59.69 & \textbf{45.42} \\
\hline
\multirow{4}{*}{MI-FGSM} & Ens & 60.23 & 43.38 & 38.20 & 35.59 & 38.40 & 47.37 & 42.79 & 50.76 & 44.59 \\
 & SVRE & 87.59 & 76.19 & 55.32 & 57.34 & 49.32 & 56.76 & 45.95 & 55.41 & 60.49 \\
 & AdaEA & 73.12 & 51.32 & 44.43 & 40.97 & 55.13 & 73.53 & 65.12 & 74.04 & 59.71 \\
 & HEAT & 69.71 & 65.26 & 61.08 & 59.51 & 59.72 & 69.16 & 51.40 & 68.69 & \textbf{63.07} \\
\hline
\multirow{4}{*}{DI$^{2}$-FGSM} & Ens & 78.42 & 56.00 & 44.13 & 41.74 & 32.83 & 43.72 & 44.10 & 55.22 & 49.52 \\
 & SVRE & 69.06 & 62.79 & 52.81 & 52.81 & 44.31 & 53.06 & 47.17 & 53.15 & 54.40 \\
 & AdaEA & 80.76 & 59.49 & 52.62 & 48.21 & 56.52 & 71.39 & 66.99 & 74.86 & 63.86 \\
 & HEAT & 75.44 & 68.28 & 63.44 & 64.16 & 61.86 & 75.51 & 60.88 & 70.65 & \textbf{67.53} \\
\hline
\end{tabular}
\caption{Attack success rate (\%) of black-box models on CIFAR-10 based on ensemble attacks with different attack methods.}
\label{tab:iter}
\end{table*}

\begin{table}[ht]
\centering
\begin{tabular}{cccccccc}
\hline
 \textbf{A} & \textbf{B} & \textbf{C} & \textbf{D} & \textbf{CNNs} & \textbf{ViTs} & \textbf{All} & \textbf{Black} \\
\hline
 & & & & 31.37 & 13.97 & 22.67 & 12.96 \\
 \checkmark & & & & 39.74 & 44.28 & 42.01 & 31.72 \\
 \checkmark & \checkmark & & & 40.20 & 50.81 & 45.50 & 36.70 \\
 \checkmark & \checkmark& \checkmark & & 41.58 & 53.90 & 47.74 & 39.10 \\
 \checkmark & \checkmark & & \checkmark & 41.13 & 53.59 & 47.36 & 38.74 \\
 \checkmark & \checkmark & \checkmark & \checkmark & 41.80 & 55.02 & 48.41 & 40.31 \\
\hline
\end{tabular}
\caption{Ablation study results for various components in HEAT.}
\label{tab:ablation}
\end{table}

\textbf{Models.} 
We utilize black-box models from both CNNs and ViTs, including ResNet-50 \cite{he2016deep}, WideResNet-50 \cite{zagoruyko2016wide}, BiT-M-R50×1 \cite{kolesnikov2020big}, and BiT-M-R101 \cite{kolesnikov2020big} for CNNs, and ViT-Base \cite{dosovitskiy2020image}, DeiT-Base \cite{touvron2021training}, Swin-Base \cite{liu2021swin}, and Swin-Small \cite{liu2021swin} for ViTs. For ensemble attacks, we use white-box models such as ResNet-18 \cite{he2016deep}, Inception v3 \cite{szegedy2016rethinking}, ViT-Tiny \cite{dosovitskiy2020image}, and DeiT-Tiny \cite{touvron2021training}.

\textbf{Implementation.}
We set the perturbation bound as $\epsilon = \frac{8}{255}$ and the common gradient contribution ratio as $p = 0.7$. For I-FGSM \cite{kurakin2018adversarial}, the number of iterations is set to 10, and the step size $\alpha = \frac{\epsilon}{10}$. For MI-FGSM \cite{dong2017discovering}, the momentum is set to 0.9. For DI$^{2}$-FGSM \cite{xie2019improving}, the resize rate and diversity probability are set to 0.9 and 0.5, respectively. The batch size for all experiments is set to 10. All experiments were conducted using Pytorch version 1.12.1 and Python version 3.8.0, and implemented on an NVIDIA GeForce RTX 4090.

\begin{table*}[ht]
\centering
\footnotesize
\begin{tabular}{ccccccccccc}
\hline
\multirow{2}{*}{Ens. Types} & \multirow{2}{*}{Ens. Models} & \multirow{2}{*}{Attack} & \multicolumn{4}{c}{CNNs} & \multicolumn{4}{c}{ViTs} \\ \cline{4-11} 
 &  &  & Res-50 & WRN101-2 & BiT-101 & Avg. & ViT-B & DeiT-B & Swin-S & Avg. \\  \hline
\multirow{4}{*}{Only CNNs} & \multirow{2}{*}{Res-18, Inc-v3} & Ens & 13.51 & 12.00 & 8.28 & 11.26 & 0.67 & 3.97 & 5.30 & 3.31 \\  

 &  & HEAT & 25.38 & 20.45 & 14.57 & \textbf{20.13} & 0.75 & 5.60 & 9.74 & \textbf{5.36} \\ \cline{2-11}
 & \multirow{2}{*}{\parbox{2cm}{Res-18, Inc-v3, BiT-50}}& Ens & 21.21 & 19.64 & 26.09 & 22.31 & 2.41 & 8.28 & 8.88 & 6.52 \\ 
 && HEAT & 26.60 & 22.71 & 34.98 & \textbf{28.10} & 5.35 & 8.10 & 17.19 & \textbf{10.21} \\ \cline{1-11}

\multirow{4}{*}{Only ViTs} & \multirow{2}{*}{ViT-T, DeiT-T} & Ens & 31.08 & 28.67 & 38.62 & 32.79 & 42.95 & 46.36 & 41.72 & 43.68 \\  

 &  & HEAT & 40.38 & 29.48 & 31.14 & \textbf{33.67} & 47.32 & 64.84 & 62.38 & \textbf{58.18} \\ \cline{2-11}
 & \multirow{2}{*}{\parbox{2cm}{ViT-T, DeiT-T, Swin-B}}& Ens & 34.43 & 30.65 & 39.67 & 34.92 & 41.46 & 43.20 & 57.60 & 47.42 \\ 
 && HEAT & 39.42 & 31.23 & 35.64 & \textbf{35.43} & 45.28 & 63.24 & 62.81 & \textbf{57.11} \\ \cline{1-11}

 \multirow{6}{*}{Only ViTs} & \multirow{2}{*}{\parbox{2cm}{Res-18, Inc-v3, ViT-T}} & Ens & 20.11 & 19.77 & 11.18 & 17.02 & 2.86 & 11.80 & 11.80 & 8.82 \\  

 &  & HEAT & 31.48 & 24.89 & 23.47 &  \textbf{26.61}&  20.09& 29.28 & 31.08 & \textbf{26.82}  \\ \cline{2-11}

 & \multirow{2}{*}{\parbox{2cm}{Inc-v3, ViT-T, DeiT-T}} & Ens & 19.91 & 13.58 & 8.84 & 14.11 & 6.09 & 10.47 & 10.59 & 9.05 \\  

 &  & HEAT & 29.22 & 23.32 & 21.74 & \textbf{24.76} & 21.02 & 31.86 & 32.59 & \textbf{28.49} \\ \cline{2-11}
 
 & \multirow{2}{*}{\parbox{2cm}{Res-18, Inc-v3, ViT-T, DeiT-T}}& Ens & 17.2 & 22.56 & 12.32 & 17.36 & 6.83 & 10.49 & 12.63 & 9.98 \\ 
 && HEAT & 39.17 & 32.58 & 35.51 & \textbf{35.75} & 38.18 & 51.57 & 51.12 & \textbf{46.96} \\ \cline{1-11}
\end{tabular}
\caption{Performance comparison of HEAT and Ens methods on different ensemble types}
\label{tab:type}
\end{table*}

\subsection{Main Result}

We compared HEAT with other ensemble attack methods on three datasets—CIFAR-10, CIFAR-100, and ImageNet—by evaluating their attack success rates ($ASR$) on eight black-box models and averaging the $ASR$. As shown in Table \ref{tab:performance}, HEAT consistently outperforms other methods, both on simple datasets like CIFAR-10 (10 classes) and complex ones like ImageNet (1000 classes). HEAT shows exceptional performance, particularly on advanced models like Swin-B and ViT-B. Compared to the baseline Ens method, HEAT achieves an average $ASR$ improvement exceeding 28\%.

\begin{figure}[ht]
\centering
\includegraphics[width=1\linewidth]{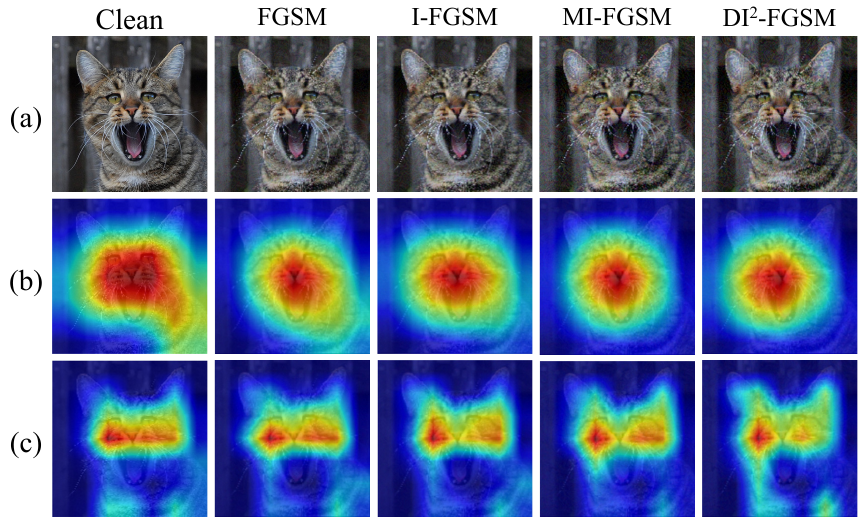}
\caption{Visualization of adversarial attacks (a) and their impact on attribution maps for Resnet-18 (b) and WideResnet-50 (c) models using Grad-CAM.}
\label{fig:attri}
\end{figure}

We also explored integrating different ensemble attack methods with base attack strategies, including I-FGSM, MI-FGSM, and DI$^{2}$-FGSM. As summarized in Table \ref{tab:iter}, HEAT achieves $ASR$ improvements of 5.11\%, 22.76\%, and 27.22\% for the three ensemble attack methods, significantly outperforming others.

\subsection{Ablation Studies}

In this section, we perform an ablation study on CIFAR-10 to assess the contribution of each component in HEAT to its overall attack effectiveness. Specifically, HEAT is decomposed into the following four key components:
\begin{compactitem}
\item A: C-GRADS; 
\item B: Intra-domain Coherence; 
\item C: Loss Contribution Factor; 
\item D: Alignment Contribution Factor. 
\end{compactitem}
We evaluate the impact of these components, both individually and combined, on the $ASR$ by incrementally incorporating them into the Ens method. As shown in Table \ref{tab:ablation}, the $ASR$ improves significantly with each component’s addition. Notably, the integration of C-GRADS significantly enhances cross-model transferability, showing exceptional performance on complex models like ViTs.

\subsection{Further Analysis}
\label{sec:further}
\textbf{Ensemble types.} We evaluated HEAT on both homogeneous ensembles (only CNNs or ViTs) and heterogeneous ensembles (mix), as shown in Table \ref{tab:type}. Our results reveal that the $ASR$ improves as the number of white-box models increases. Interestingly, ViT models demonstrate a significant advantage in adversarial attacks, whereas CNN models offer more balanced performance. 

\textbf{Attribution Performance.} To illustrate attack performance intuitively, we generated adversarial examples with various attack methods and visualized their attribution maps for both white-box (ResNet-18) and black-box (WideResNet-50) models using Grad-CAM, as shown in Figure \ref{fig:attri}. In Figure \ref{fig:attri}(a), the adversarial examples are visually indistinguishable from the originals to humans but cause incorrect model predictions. Figures \ref{fig:attri}(b) and \ref{fig:attri}(c) demonstrate that as attack intensity increases, the regions in the attribution maps associated with the ground truth progressively shrink, ultimately leading to model misclassification.

\section{Conclusion}

In this work, we introduce HEAT, a novel method that enhances adversarial transferability by incorporating domain generalization principles into model ensemble attacks. HEAT uses its two core modules, C-GRADS and D-HARMO, to synthesize shared gradient directions and dynamically optimize gradient contributions for highly transferable adversarial perturbations. By redefining ensemble attack strategies through domain generalization, HEAT addresses challenges in gradient alignment and adaptive weighting, improving attack success rates on black-box models. Extensive experiments show HEAT’s superiority over existing methods, establishing it as a robust solution for enhancing adversarial transferability.




\section*{Impact Statement}

This paper presents work whose goal is to advance the field of  Machine Learning. There are many potential societal consequences  of our work, none which we feel must be specifically highlighted here.

\nocite{langley00}

\bibliography{example_paper}
\bibliographystyle{icml2025}

\newpage
\appendix
\onecolumn


\end{document}